# Rule-based query answering method for a knowledge base of economic crimes

Jaroslaw Bak

Institute of Control and Information Engineering,
Poznan University of Technology,
M. Sklodowskiej-Curie Sqr. 5, 60-965 Poznan, Poland
Jaroslaw.Bak@put.poznan.pl

**Abstract.** We present a description of the PhD thesis which aims to propose a rule-based query answering method for relational data. In this approach we use an additional knowledge which is represented as a set of rules and describes the source data at concept (ontological) level. Queries are posed in the terms of abstract level. We present two methods. The first one uses hybrid reasoning and the second one exploits only forward chaining. These two methods are demonstrated by the prototypical implementation of the system coupled with the Jess engine. Tests are performed on the knowledge base of the selected economic crimes: fraudulent disbursement and money laundering.

## 1 Introduction

Data stored in relational databases are described only by their schema (syntactic structure of data). Therefore, it is often difficult to pose a query at a higher level of abstraction than in a language of database relations and attributes. There is also a mismatching problem with table and column names without strictly defined semantics. A lack of a conceptual knowledge can be overcome by introducing ontologies which for evaluation purposes can be transformed into a set of rules. This kind of additional rule-based knowledge allows reasoning and query answering at an appropriate abstract level and relieves a user of using structural constructions from SQL. This kind of query evaluation is called a rule-based query answering method.

In our rule-based system we apply rules that are Horn clauses [8]. If there is conjunction of several predicates in the head, the rule can be easily transformed into Horn clauses with the Lloyd-Topor transformation [8].

We assume that only unary or binary predicates exist in our system, according to the terms that appear in OWL language [2] (since we decided to use this standard as a way to express conceptual knowledge).

Every rule consists of two parts: the left-hand-side, which is called the body, and the right-hand-side, which is called the head. Generally both parts are the sets of atoms that are interpreted conjunctively. In the body of the rule we use premises (patterns, conditions), which have to be satisfied by the appropriate atoms (facts) to allow the rule to be fired and produce conclusions from the rule's head. Next section describes the problem statement of the proposed PhD thesis. Section 3 presents current knowledge of the problem domain and existing solutions. Section 4 contains results achieved so far, the current state of the work and author's contributions. In Section 5 the concluding remarks are given.

## 2 Problem statement

The presented PhD thesis is trying to cope with the following research question:

*How to efficiently query a relational database*
*at the conceptual level defined in a rule-based system?*

This question is strongly connected with the following three main problems:

i. Rule-based query answering,
ii. The combination of a rule-based system and a relational database,
iii. The construction of the knowledge base (i.e. knowledge of economic crimes).

In a rule-based query answering method we assume that there exists a knowledge base which contains two parts: intensional and extensional. The intensional knowledge is represented as a set of rules and describes the source data at a conceptual (ontological) level. The extensional knowledge consists of facts that are stored in the relational database as well as facts that were derived in the reasoning process. Queries can be posed in the terms of the conceptual level. Thus, one gets an easier way to create a query than using structural constructions from SQL (Structured Query Language). The rule-based query answering method uses the reasoning process to obtain an answer for a given query. During this process facts from database are gathered and used to derive new facts according to a given set of rules. Next, the answer is constructed and presented.

In the first two problems (i, ii), we need to deal with the following questions:

1. What kind of rule-based system do we want to use?
2. How to express and represent the conceptual knowledge in the form of rules?
3. What is the language of the queries that can be evaluated by the system?
4. What kind of reasoning is involved in the rule-based query answering?
5. How to ensure the decidability of the query answering method?
6. How to combine a relational database with a rule-based system?
7. Which reasoning engine should be used for the prototypical implementation?
8. What are other potential applications of the proposed method and system?

We also assume that the rule-based query answering method will be used with the knowledge base of the selected economic crimes: fraudulent disbursement and money laundering. Particularly, we assume our system to be aimed at determining legal sanctions for crime perpetrators and to discover crime activities and roles (of particular types of owners, managers, directors and chairmen) using concepts, appropriate relations and rules.

The answers to the majority of the given questions and current achievements are presented in Section 4.

## 3   Overview of existing solutions

The presented problem, the rule-based query answering task [14], has many times been subject to research. Generally, there are two kinds of reasoning method applied in the rule-based query answering task. The first one is a backward chaining method, where reasoning is goal-driven. In this case our goal is the query posed to the system. This scheme of reasoning is implemented, for instance, in Prolog engine, and takes the form of the Selective Linear Definite clause resolution (SLD). In the backward reasoning technique facts are obtained only when they are needed in derivations.

On the contrary a forward chaining approach, which is data-driven, needs reasoning about all facts. In the working memory some of the inferred facts are useless and many rules are fired unnecessarily. It has a negative impact on the efficiency of the answering process. Moreover, because all facts should exist in the working memory, the scalability of reasoning task is poor due to the limited RAM memory. This drawback occurs also in the backward chaining.

The rule-based query answering task in rule-based systems, which exploits forward chaining is generally an inefficient method. The results of the OpenRuleBench initiative [1] show that efficiency of tabling Prolog and deductive database technologies surpasses the ones obtained from the corresponding pure rule-based forward chaining engines.

The most comprehensive approaches concerning optimizations of bottom-up query evaluation (in forward chaining) were given in [14, 15]. The general method relies on the transformation of a program $P$ (set of rules) and a query $Q$ into a new program, *magic(P $\cup$ Q)*, as shown in [15]. This *magic transformation* modifies each original rule by additional predicates to ensure that the rule will fire only when the values for these predicates are available. There were also other improvements and modifications of magic approach [14]. According to the work presented in [12] we also believe that the bottom-up approach has still room for improvements in order to increase the performance of the rule-based query answering task.

There exist also some works about the combination of rules (or logic programming) with relational databases. Notable are approaches presented in [18], [10] and [19] where ontology-based data access is performed with Prolog rules or Disjunctive Datalog.

The problem of applying rules in economic crimes is quite new. Most of the research work in the legal area relies on using ontologies in the field of information management and exchange [23, 24], not reasoning [16]. Other solutions, developed for instance in FFPoirot project [25, 26], concern descriptions of financial frauds, mainly the Nigerian letter fraud and fraudulent Internet investment pages. The ontologies developed in this project are not publicly available.

In our approach rules and queries are used to reflect data concerning documents and their attributes, formal hierarchy in a company, parameters of transactions, engaged people actions and their legal qualifications. To the best of our knowledge it is the first such approach in the field of economic crimes, besides the work presented in [17], which concerns cybercrimes.

## 4 Achievements and the current work

### 4.1 General assumptions

As mentioned in Section 2, the first two tasks include eight questions. For most of them, in the current state of our work, the answers are already known:

1. We wanted to use a production rule system because we need to apply our solutions in the real world applications.
2. We decided to express the conceptual knowledge with the Horn-SHIQ ontology combined with SWRL (Horn-like) [3] rules. We are aware of the restrictions that Horn-SHIQ imposes on ontology creation [21], but this fragment of OWL is sufficient for our needs.
3. Currently we assume conjunctive queries only, which are built of the terms from ontology (concepts and relations).
4. We developed two ways of applying reasoning process in the rule-based query answering task. In the first one [6] *hybrid* reasoning (forward and backward chaining) is used and in the second one only *forward* chaining and *extended* rules are executed. The second approach is still in progress.
5. We use the *Datalog Safety* [11] restriction in the rule-based query answering and *DL-safe* rules in ontology creation [22].
6. We developed the special mapping method which is presented later in this section and also in [6].
7. We decided to use the *Jess* (Java Expert System Shell) engine [4, 5], since it is one of the fastest commercial engines (with the free academic use) and it can be easily integrated with the Java language (which is the implementation language of our tool). The Jess engine also supports both forward and backward chaining.
8. We are convinced that our knowledge base of economic crimes [27, 30] would not be the only application of the defined system. Our methods are general and can be used in every application, which requires additional knowledge for query evaluation or need to offer an easier way of query creation than with the traditional SQL.

Our current results were presented in Polish and English papers [6, 27, 28, 29, 30].

### 4.2 Query answering with the hybrid reasoning

The approach described in [6] concerns the hybrid reasoning in the rule-based query answering task. In this work we described also the method of mapping between an ontology and a relational database. We presented our prototypical implementation of a library tool, the Semantic Data Library (SDL), which integrates the Jess engine, rules and ontology to effectively query a relational database.

In our hybrid reasoning process the backward chaining engine is responsible only for gathering data from a relational database. Data is added (asserted in Jess terminology) as triples into the working memory. The forward chaining engine can answer a query with all constraints put on variables in a given query (=, !=, <, > etc.).

The queries are constructed in Jess language in terms of ontology concepts. The mapping between the ontology classes and properties and the relational database schema is defined to fit syntactic structures and to preserve the semantics of the data.

Extensional data itself is stored in a relational database. The ontology and the mapping rules transformed into Jess language format provide the additional semantic layer to the relational database. Such an approach allows for answering queries to a relational database with a reasoning process performed in the Jess system over rules and ontology. The hybrid reasoning and query execution is supported by the SDL library. More details are given in [6].

### 4.3 Mapping between ontology terms and relational database

A mapping between ontology terms and relational data [6] is defined as a set of rules where each rule is of the following form:

*SQL query => essential predicate*

where "essential" means that the instance of the term cannot be derived from the rules. We assume that every "essential" ontology term has its appropriate SQL query and can be obtained only in a direct way, as a result of the SQL query. For example, in the following OWL hierarchy of classes *Mother is-a Woman is-a Person*, where the class *Mother* is a subclass of the class *Woman* etc., every instance of the class *Mother* is an "essential" term.

We assume that every SQL query has the following form:

*SELECT [R] FROM [T] <WHERE> <C, AND, OR>*

where:

- *R* denotes the result attributes (columns) – one or two according to the unary or binary terms (OWL Class, OWL DataProperty or OWL ObjectProperty),
- *T* stands for the tables, which are queried,
- *WHERE* is an optional clause to specify the constraints,
- *C* abbreviates the constraints in the following form: <attribute, comparator, value>, for example: *Age > 21*,
- *AND, OR* are the optional SQL commands.

As an example, let us assume that we have a table *Person* with the following attributes: *Id*, *Name*, *Age* and *Gender*. To obtain all adult men, we would define the following SQL query: *SELECT Id FROM Person WHERE Age>21 AND Gender='Male'*.

The mapping process requires defining SQL queries for all "essential" classes and properties. Other terms can be mapped too, but this is not necessary, since instances of them can be derived in the reasoning process.

### 4.4 Knowledge base of economic crimes

The approaches presented in [27, 28, 29] concern construction of the knowledge base of the selected economic crimes: fraudulent disbursement and money laundering. We analysed current related works and proposed the formal model of these economic crimes. We developed the ontology, which is the result of an analysis of about 10 crime cases. This means that the ontology is crafted to a task rather than attempting to describe the whole conceivable space of concepts and relations (top ontologies). The intensional part of the knowledge base contains also SWRL rules, which are very important when we want to determine legal sanctions for crime perpetrators and to discover crime activities and roles (not only to describe a crime scheme).

The methodology consists of several steps:

1. Design of a hierarchical data representation with 'minimal' ontology, which is used to uncover a crime scheme. This means using only necessary concepts that follow in the logical order of uncovering a crime. In the first stage goods/services transfer data is analyzed with relation to 3 basic flows: money, invoices, and documents (i.e., confirming that the service or goods have been delivered). In addition, responsible or relevant people within companies are associated with particular illegal activities.
2. Provision of a framework in which the graph building process and queries are executed.
3. Relating answers to queries with crime qualifications.

This approach is limited, but provides an essential model for evidence-building of a very important class of financial crimes: among them acting to do damage to a company and money laundering. Both crimes occurred in the example *Hydra Case* which was tested with the hybrid approach and artificially generated data. The work and results are presented in [30].

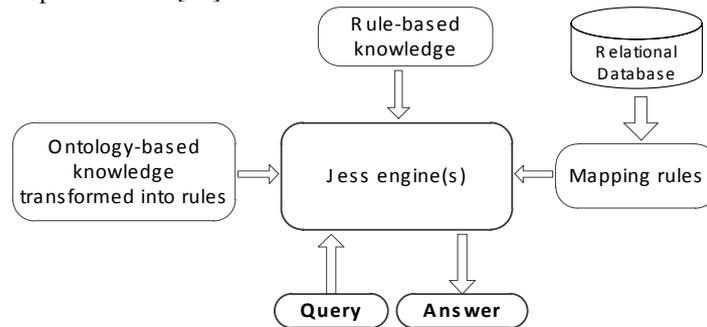

Figure 1. The architecture of the rule-based query answering system

### 4.5 Current work

In the current state of our work we are focused on the new rule-based query answering method which uses *extended* rules. *Extended* means that these rules are generated automatically from the basic ones for the evaluation purposes, and the modification is strongly connected with the magic transformation method. The extended rules are generated in the goal- and dependency-directed transformation. In this method we are interested in dependencies between variables appearing in predicates inside each rule.

The rule-based query answering method in this approach needs the different assumptions from the hybrid one because we use only one Jess engine to obtain relational data and answer a query. Obviously, we have to modify our query answering algorithm prepared for the hybrid system. This work is still in progress and results will be presented as soon as possible.

Figure 1 presents the architecture of our system which covers both solutions (hybrid reasoning and reasoning with extended rules).

## 5   Conclusions

In this paper we have outlined the content of the PhD thesis titled *Rule-based query answering method for a knowledge base of economic crimes*. Up to date we have obtained some achievements in the research, particularly related to the special crime-oriented ontological knowledge, its representation in rules of the Jess system, the connection with extensional data in a database and query answering by reasoning over the different data representations. We continue our research aiming to elaborate a new method of rules transformation, which will allow for more efficient application of rules in query answering task. We have to manage with problems presented in Section 2 and to provide a precise, clear and formal description of our solutions. We have already obtained positive results of tests performed on the prototype system but we also plan to execute queries prepared by the OpenRuleBench initiative. The comparison of our results and those obtained in a pure Jess system seems to be an adequate and objective assessment of usefulness of our work.

## References

1. Liang S., Fodor P., Wan H., Kifer M., OpenRuleBench: An Analysis of the Performance of Rule Engines, Proceedings of the 18th international conference on World wide web, ACM, 2009, o. 601-610.
2. McGuinness D., van Harmelen, F.: Owl web ontology language overview. W3C Recommendation, 10 February 2004, http://www.w3.org/TR/owl-features/
3. Horrocks I., Patel-Schneider, P.F., Boley, H., Tabet, S., Grosof, B., Dean, M.: Swrl: A semantic web rule language combining owl and ruleml. W3C Member Submission (May 21 2004), http://www.w3.org/Submission/SWRL/
4. Jess (Java Expert System Shell), http://jessrules.com/
5. Friedman-Hill, E.: Jess in Action, Manning Publications Co. (2003)
6. Bak, J., Jedrzejek, C., Falkowski, M.: Usage of the Jess engine, rules and ontology to query a relational database. In: Governatori, G., Hall, J., Paschke, A. (eds.) RuleML 2009. LNCS, vol. 5858, pp. 216–230. Springer, Heidelberg (2009)
7. Forgy C., Rete: A Fast Algorithm for the Many Pattern/Many Object Pattern Match Problem, Artificial Intelligence, 19, pp. 17-37, 1982.
8. Lloyd J.W., Foundations of logic programming (second, extended edition). Springer series in symbolic computation. Springer-Verlag, New York, 1987.
9. Horridge M., Bechhofer S. The OWL API: A Java API for Working with OWL 2 Ontologies. OWLED 2009, 6th OWL Experienced and Directions Workshop, Chantilly, Virginia, October 2009.
10. Poggi A., Lembo D., Calvanese D., De Giacomo G., Lenzerini M., Rosati R., Linking Data to Ontologies, Journal on Data Semantics, Volume 10, 2008, p. 133-173.

11. Gallaire H., Minker J. (Eds.): Logic and Data Bases, Symposium on Logic and Data Bases, Centre d'études et de recherches de Toulouse, 1977. Advances in Data Base Theory, Plenum Press, New York, 1978
12. Brass, S., Implementation Alternatives for Bottom-Up Evaluation, 26th International Conference on Logic Programming, ICLP (Technical Communications), Edinburgh 2010, 44-54.
13. Liang S., Fodor P., Wan H., Kifer M., OpenRuleBench: Detailed Report, May 29, 2009, http://projects.semwebcentral.org/docman/view.php/158/69/report.pdf.
14. F. Bry, N. Eisinger, T. Eiter, T. Furche, G. Gottlob, C. Ley, B. Linse, R. Pichler, and F. Wei. Foundations of Rule-Based Query Answering. In Proceedings of Summer School Reasoning Web 2007, Dresden, Germany (3rd–7th September 2007), volume 4634 of LNCS, pages 1–153. REWERSE, 2007.
15. Beeri C., Ramakrishnan R., On the power of magic, Journal of Logic Programming, v.10 n.3-4, p.255-299, April/May 1991.
16. Breuker, J. (2009). Dreams, awakenings and paradoxes of ontologies, invited talk presentation, 3rd Workshop on Legal Ontologies and Artificial Intelligence Techniques, http://ontobra.comp.ime.eb.br/apresentacoes/keynoteontobra-2009.ppt.
17. Bezzazi, H. (2007). Building an ontology that helps identify articles that apply to a cybercrime case, Proceedings of the Second International Conference on Software and Data Technologies, ICSOFT 2007, Barcelona, Spain, pp. 179–185.
18. Lukácsy G., Szeredi P., Scalable Web Reasoning Using Logic Programming Techniques. In A. Polleres and T. Swift, editors, Proceedings of the Third International Conference on Web Reasoning and Rule Systems, RR 2009, Chantilly, VA, USA, October 25-26, 2009, volume 5837 of Lecture Notes in Computer Science, pages 102{117. Springer, 2009.
19. U. Hustadt, B. Motik, U. Sattler. Reducing SHIQ-Description Logic to Disjunctive Datalog Programs. Proc. of the 9th International Conference on Knowledge Representation and Reasoning (KR2004), June 2004, Whistler, Canada, pp. 152-162.
20. Hustand U., Motik B., Sattler U., Data Complexity in Very Expressive Description Logics, In Proc. of the 19th Joint Int. Conf. on Artificial Intelligence (IJCAI 2005), 2005.
21. Horn-SHIQ restrictions: http://www.w3.org/2007/OWL/wiki/Tractable_Fragments#5
22. Motik B., Sattler U., and Studer R., Query answering for owl-dl with rules. Journal of Web Semantics: Science, Services and Agents on the World Wide Web, 3(1):41–60, 2005.
23. Biasiotti, M., Francesconi, E., Palmirani, M., Sartor, G. and Vitali, F. (2008). Legal Informatics and Management of Legislative Documents. Global Centre for ICT in Parliament, Working Paper No. 2, United Nations, Department of Economics and Social Affairs.
24. Casellas, N. (2008). Modelling Legal Knowledge through Ontologies. OPJK: the Ontology of Professional Judicial Knowledge, PhD thesis, Universitat Autónoma de Barcelona, Barcelona.
25. Kerremans, K. and Zhao, G. (2005). Topical ontology of VAT, Technical Report of the FFPOIROT IP project (IST-2001- 38248). Deliverable D2.3 (WP 2), STARLab VUB.
26. Zhao, G. and Leary, R. (2005). AKEM: an ontology engineering methodology in FFPoirot, Technical Report of the FFPOIROT IP project (IST-2001-38248). Deliverable D6.8 (WP 6), STARLab VUB.
27. Bak, J., Jedrzejek, C., Falkowski, M.: Application of an Ontology-Based and Rule-Based Model to Selected Economic Crimes: Fraudulent Disbursement and Money Laundering, Lecture Notes in Computer Science, 2010, Volume 6403/2010, 210-224
28. Bak J., Jedrzejek C., Application of an Ontology-based Model to a Selected Fraudulent Disbursement Economic Crime. In P. Casanovas, U. Pagallo, G. Ajani, and G. Sartor, editors, AI approaches to the complexity of legal systems, LNAI, Vol 6237, 2010
29. Jedrzejek C., Cybulka J., Bak J., Towards Ontology of Fraudulent Disbursement, Agent and Multi-Agent Systems: Technology and Applications, Proceedings of 5th International Conference, LNCS 2011, to be published.
30. Bak J., Jedrzejek C., Falkowski M., Application of the SDL Library to Reveal Legal Sanctions for Crime Perpetrators in Selected Economic Crimes: Fraudulent Disbursement and Money Laundering, In Proceedings of the 4th International RuleML-2010 Challenge, Washington, DC, USA, October, 21-23, 2010. Edited by: Palmirani M., Omair Shafiq M., Francesconi E., Vitali F., Volume 649.